\documentclass{article}
\PassOptionsToPackage{numbers, sort&compress}{natbib}
 \usepackage[dblblindworkshop, final]{neurips_2026}

\usepackage[utf8]{inputenc} 
\usepackage[T1]{fontenc}    
\usepackage{hyperref}       
\usepackage{url}            
\usepackage{booktabs}       
\usepackage{amsfonts}       
\usepackage{nicefrac}       
\usepackage{microtype}      
\usepackage{xcolor}         
\usepackage{float}

\usepackage{amsmath}
\usepackage{graphicx}
\usepackage{multirow}
\usepackage{blindtext}
\usepackage{graphicx}
\usepackage{amssymb}
\usepackage{pifont}
\usepackage{xcolor}
\usepackage{tikz}
\usetikzlibrary{matrix,arrows.meta}

\newcommand{\cmark}{\textcolor{green!50!black}{\ding{51}}} 
\newcommand{\xmark}{\textcolor{red!70!black}{\ding{55}}}   
\newcommand{\pmark}{\textcolor{gray!70!black}{±}}   

\usepackage{pifont}
\usepackage{tabularx}

\usepackage[capitalize]{cleveref}

\title{Naga: Vedic Encoding for Deep State Space Models}

\author{
  Melanie Schaller \\
  Leibniz Universität Hannover \\
  \texttt{schaller@tnt.uni-hannover.de}
  \And
  Nick Janssen \\
  Leibniz Universität Hannover \\
  \texttt{janssen@tnt.uni-hannover.de}
  \And
  Bodo Rosenhahn \\
  Leibniz Universität Hannover \\
  \texttt{rosenhahn@tnt.uni-hannover.de}
}

\begin{document}

\maketitle

\begin{abstract}
This paper presents Naga, a deep State Space Model (SSM) encoding approach inspired by structural concepts from Vedic mathematics. The proposed method introduces a bidirectional representation for time series by jointly processing forward and time-reversed input sequences. These representations are then combined through an element-wise (Hadamard) interaction, resulting in a Vedic-inspired encoding that enhances the model’s ability to capture temporal dependencies across distant time steps. We evaluate Naga on multiple long-term time series forecasting (LTSF) benchmarks, including ETTh1, ETTh2, ETTm1, ETTm2, Weather, Traffic, and ILI. The experimental results show that Naga outperforms 28 current state of the art models and demonstrates improved efficiency compared to existing deep SSM-based approaches. The findings suggest that incorporating structured, Vedic-inspired decomposition can provide an interpretable and computationally efficient alternative for long-range sequence modeling.
\end{abstract}

\section{Introduction}
Expressive representations of time series are essential for forecasting models that must capture long-range temporal dependencies across domains such as finance, energy systems, and climate dynamics. In this work, we propose Naga — a novel architecture for deep state space models (SSMs) that integrates principles of Vedic mathematics to impose structured temporal interactions, thereby improving predictive performance.

The initial inspiration for the Vedic encoding stemmed from the concept of Vedic mathematics, specifically the sutras that describe efficient ways to multiply numbers by decomposing them into smaller, interacting components. In traditional Vedic multiplication, partial products are computed along diagonals and then combined to form the final result—a process that inherently captures structured interactions between digits. Translating this idea to neural representations, the original intention was to distribute the matrix operations in a similar way: to conceptually decompose the linear projections into structured submatrices, making it easier to analyze what parts of the model contributed positively or negatively to its performance.

However, through experimentation, we learned that the Vedic embedding itself, without explicit matrix subdivision, can be used to improve the performance of Deep SSM Cells like the Mamba2 Cell~\cite{Gu2024Mamba2} to predict long time-series sequences. By coupling each time index $t$ with its symmetric counterpart $T - t + 1$, the encoding introduces a form of bilinear interaction that naturally captures long-range dependencies in the data. This not only enhanced the interpretability of the model’s internal representations but also significantly improved its predictive performance compared to the 28 benchmarking models from the related work section on the seven most frequently used benchmark datasets. Thus, while the approach began as a conceptual borrowing from ancient mathematical principles, it ultimately evolved into a modern architectural feature that bridges symbolic structure and deep learning dynamics in a surprisingly effective way.

\begin{figure}[t]
    \centering
    \includegraphics[width=0.75\textwidth]{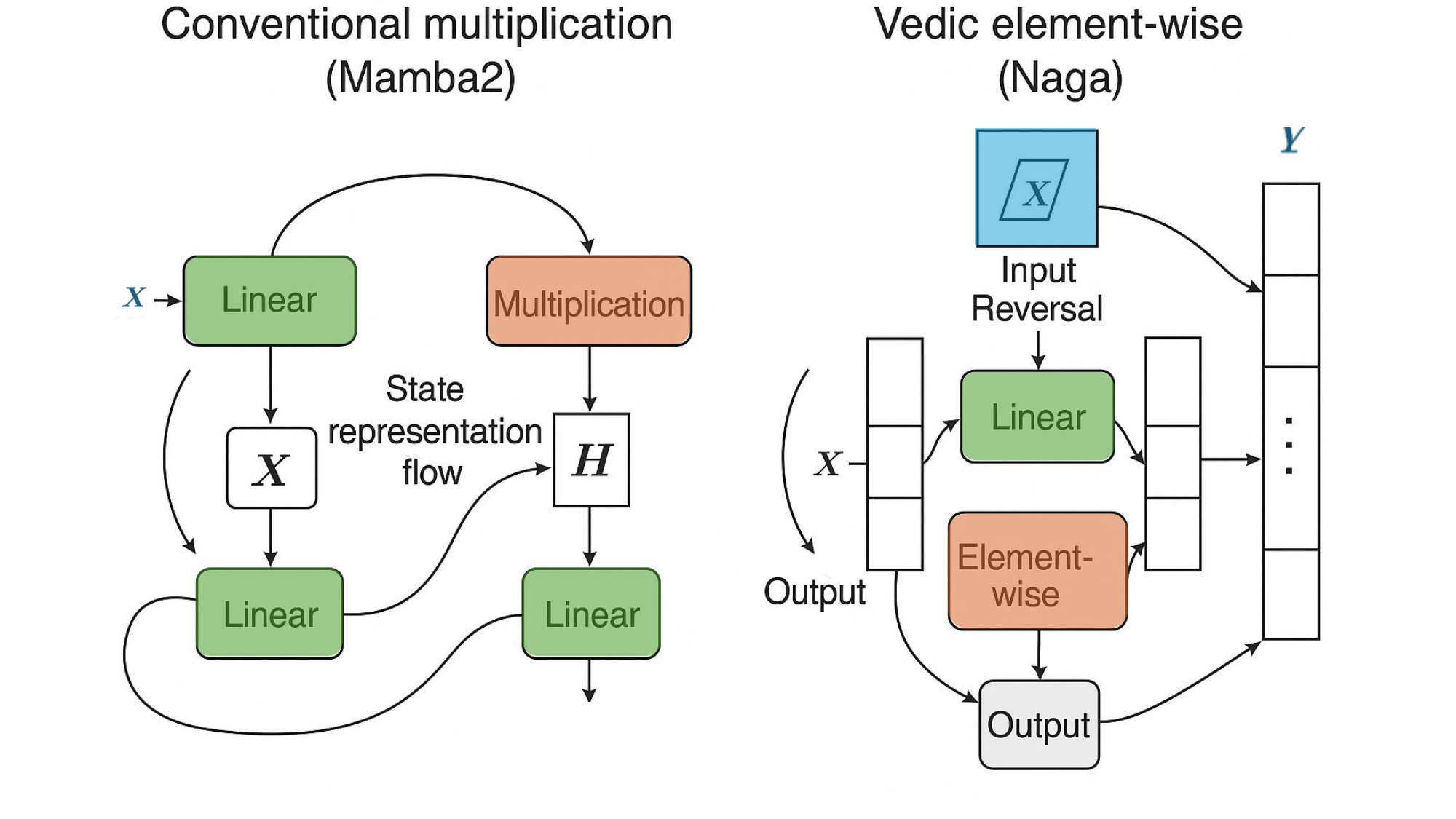}
    \caption{Comparison between conventional multiplication in Mamba2~\cite{Gu2024Mamba2}; (left) and the proposed Vedic element-wise decomposition in Naga (right). 
    While Mamba2~\cite{Gu2024Mamba2} performs standard matrix multiplications for state updates, Naga introduces bidirectional input reversal and element-wise operations inspired by Vedic mathematics. 
    This decomposition enhances gradient flow and enables the model to capture complex temporal dependencies more efficiently.}
    \label{fig:naga_comparison}
\end{figure}

Figure~\ref{fig:naga_comparison} visually illustrates the concept behind Naga, highlighting the distinction between conventional multiplication in Mamba2~\cite{Gu2024Mamba2} and the Vedic-inspired element-wise decomposition used in Naga. The traditional Mamba2 architecture utilizes global matrix multiplication: the input vector $\mathbf{x}$ is linearly projected and multiplied by a learned matrix $\mathbf{X}$, resulting in the hidden representation $\mathbf{H}$. On the other hand, our proposed Naga architecture employs a Vedic-inspired element-wise approach. The input $\mathbf{x}$ is first reversed in time, linearly projected, and subsequently processed through parallel Hadamard (element-wise) operations, which produces the output $\mathbf{y}$. 

\paragraph{Naga Claim:} The Vedic decomposition of multiplications as an element-wise operation in the Naga architecture facilitates deep SSMs in learning complex temporal interactions, thereby improving forecasting accuracy compared to conventional, non-decomposed representations. 

Our contributions are as follows:
\begin{itemize}
    \item We introduce the Naga model, which integrates Vedic mathematical principles into the existing framework of Mamba2~\cite{Gu2024Mamba2} deep SSMs.
    \item We experimentally prove that, the Vedic decomposition of multiplications facilitates deep SSMs in learning long time-series forecasting (for more details see also Appendix A.4-A.7).
    \item We mathematically demonstrate that the Vedic decomposition introduces an inductive bias that enhances gradient flow and representation disentanglement within deep state space models (see Appendix A.1-A.2).
    \item Across seven commonly used LTSF benchmark datasets and different sequence lengths, Naga outperforms state-of-the-art methods in terms of MSE.
    \item We make our code available on GitHub\footnote{After acceptance notification.}.
\end{itemize}



\section{Related Work}\label{sec:related_work}
\subsection{Long Time-Series Forecasting}
Given a time series $\{x_t\}_{t=1}^T$ where $x_t \in \mathbb{R}^n$ denotes the multivariate observation vector at time $t$, the objective of long time-series forecasting (LTSF) is to learn a function
\[
\hat{X}_{T+1:T+h} = f(x_{T-d+1}, x_{T-d+2}, \dots, x_T; \theta),
\]
so that $\hat{X}_{T+1:T+h} \in \mathbb{R}^{h \times n}$ represents the predicted future multivariate sequence of length $h$. Here, $d$ is the look-back window size (the number of past time steps used for prediction), $h$ is the forecasting horizon (the number of future time steps to predict), and $\theta$ denotes the learnable parameters of the model.

A comprehensive survey on long sequence time-series forecasting can be found in \citep{chen2023long}. As mentioned in the introduction, the goal of LTSF is to minimize the prediction error, which can be quantified by metrics like Mean Squared Error (MSE). The equation is given in Section~\ref{sec:model_architecture}.

\subsection{Benchmarking models}

A wide range of models has been proposed for LTSF tasks, spanning from classical recurrent and convolutional approaches to recent transformer-based architectures and linear state-of-the-art models.

Early deep learning models such as long-short-term memory networks (LSTM)~\citep{hochreiter1997long} and temporal convolutional networks (TCN)~\citep{lea2017temporal} have demonstrated strong sequence modeling capabilities. LSTNet~\citep{lai2018modeling} builds upon these by combining Convolutional Neural Networks (CNNs) for local dependencies, recurrent neural networks (RNNs) for long-term structure, and a skip-recurrent component to model periodicity. Similarly, SegRNN~\citep{lin2023segrnn} was introduced to segment input sequences and process them with hierarchical RNNs, improving interpretability and scalability.

The advent of attention-based models has shifted the paradigm. The Transformer architecture, originally developed for Natural Language Processing (NLP), was adapted for time series, though its quadratic complexity is a known bottleneck. Furthermore, transformers are inherently position-invariant, treating all input positions equally except for added positional encodings, which works well for language where order can be inferred from syntax and semantics, but poses a challenge for time series data where the exact position and sequence of values are crucial due to underlying temporal, physical, or seasonal dependencies~\citep{zhang2022detrendattendrethinkingattention}. To address this, various alternatives have emerged: Informer~\citep{zhou2021informer} uses ProbSparse attention and a self-attention distillation mechanism, enabling forecasting over extremely long sequences. LogTrans~\citep{nie2022logtrans} applies log-sparse self-attention, while Reformer~\citep{kitaev2020reformer} incorporates locality-sensitive hashing to reduce complexity. Autoformer~\citep{wu2021autoformer} further innovates with series decomposition and an autocorrelation mechanism, outperforming many earlier models on multivariate LTSF benchmarks. PRformer~\citep{yu2025prformer} combines parameter reusability with dynamic decomposition to enhance forecasting efficiency.

More recent models have explored patch-based mechanisms: PatchTST~\citep{nie2023patchtst}, PatchMixer~\citep{gong2023patchmixer}, and xPatch~\citep{stitsyuk2025xpatch} break down time series into patches, similar to techniques from vision transformers to capture temporal locality. TiDE~\citep{das2023tide} adopts an multi-layer perceptron~\citep{popescu2009multilayer} (MLP)-based architecture with time-distributed encoding for fast and accurate forecasting, and SCINet~\citep{liu2022scinet} introduces recursive splitting and fusion to extract multi-scale temporal patterns.

In parallel, linear models have gained attention due to their simplicity and strong performance. DLinear, NLinear~\citep{li2023linear}, and RLinear~\citep{li2023revisitinglongtermtimeseries}, are representative of this trend, directly modeling trend and seasonality components through linear projections. Extensions like MoLE-DLinear and MoLE-RLinear incorporate mixture-of-experts (MoE)~\citep{ni2024mixture} modules to allow for specialized sub-models, while DiPE-Linear~\citep{zhao2024dipe} combines linear forecasting with disentangled periodic encodings.

\sloppy Other hybrid or ensemble methods also contribute to the benchmarking landscape. LTBoost~\citep{truchan2024ltboost} is a boosting-based approach tailored for long-term dependencies, FiLM~\citep{zhou2022filmfrequencyimprovedlegendre} applies feature-wise linear modulation for adaptive conditioning, TTM~\citep{ekambaram2024tiny} combines multiple time series transformations for robustness, and QuerySelector~\citep{klimek2021long} introduces a mechanism to select relevant past queries dynamically.

Collectively, these models represent a comprehensive benchmarking ecosystem, ranging from interpretable linear baselines to deep transformer-based solutions and thus provide the foundation for comparing the performance and generalizability of LTSF architectures.

\subsection{Deep SSM SOTA models for time-series}

State Space Models (SSMs) have recently gained significant attention as alternatives to attention-based architectures in sequence modeling tasks. 
An SSM models the evolution of hidden states $\mathbf{h}_t$ over time according to the recursive formulation
\[
\mathbf{h}_{t+1} = A \mathbf{h}_t + B \mathbf{x}_t, \quad \mathbf{y}_t = C \mathbf{h}_t + D \mathbf{x}_t,
\]
where $A, B, C, D$ are learnable transition, input, output, and skip matrices.
This formulation allows efficient recurrent processing and continuous-time modeling, making SSMs particularly suitable for long sequences while maintaining compactness and low parameter counts~\citep{Gu2021S4}.
However, SSMs are often difficult to train due to gradient instabilities and the challenge of balancing long- and short-term dynamics.

Built upon the original Mamba architecture, S-Mamba~\citep{wang2024smamba} proposed selective SSM-kernels with dynamic token-wise adaptability, whereas Mamba-2~\citep{dao2024transformersssmsgeneralizedmodels} advanced this by integrating SSD-matrices (Semi-Separable Design) and attention-aware mechanisms to enhance temporal and cross-channel interactions. 

In contrast, our proposed Naga approach avoids complex kernel parametrizations and instead focuses on improving the internal representation of temporal relationships through a novel encoding principle inspired by Vedic mathematics. Vedic multiplication, originating from ancient Indian arithmetic systems, decomposes a product into a set of parallel, element-wise operations (often referred to as the “vertically and crosswise” method). Rather than performing dense matrix products as in conventional linear algebra, it computes structured partial products that can be aggregated with minimal interaction across components. 

Integrating this mechanism into the Mamba2~\cite{Gu2024Mamba2} state update introduces a form of \emph{inductive bias}—that is, a deliberate structural constraint guiding the model toward hypotheses that better reflect assumed properties of the data (see Appendix A.3). In our case, the bias favors interpretable temporal dependencies instead of unrestricted, fully entangled dynamics. Such biases are desirable in long-horizon forecasting because they improve generalization under limited or noisy data and make learned state transitions more transparent.

\begin{table}[ht]
\centering
\small
\renewcommand{\arraystretch}{1.2}

\begin{tabularx}{\textwidth}{p{2.2cm} X X X X}
\toprule
\textbf{Criterion} &
\textbf{Naga (Ours)} &
\textbf{S-Mamba~\citep{wang2024smamba}} &
\textbf{LTSMamba} &
\textbf{Mamba-2~\citep{dao2024transformersssmsgeneralizedmodels}} \\
\midrule

\textbf{Cell Structure} &
Mamba2-like block applied at feature level using Vedic decomposition. &
Mamba Block with parametrized SSM-kernels, bidirectional. &
Mamba2-Cell with SSD matrices, TDB, ICB. &
SSD-based Mamba2Cell with semi-separable attention kernel. \\

\textbf{Selectivity} &
\cmark~Adaptive, token-wise feature selection via element-wise decomposition. &
\cmark~Dynamic kernel selection, token-wise adaptive. &
\pmark~Gated forget mechanism + adaptive channel re-ranking. &
\cmark~Grouped-value attention with selective SSD-state spaces. \\

\textbf{Bidirectionality} &
\cmark~Bidirectional Vedic encoding. &
\cmark~Bidirectional Mamba Blocks. &
\cmark~Bidirectional TDB and ICB structures with attention. &
\pmark~Theoretically bidirectional (configurable). \\

\textbf{Inter-Channel Awareness} &
\pmark~Moderate – features processed individually, easing interaction learning. &
\xmark~Weak (standard multivariate input). &
\cmark~Strong via ICB and adaptive re-ranking. &
\pmark~Medium (grouped-value attention). \\

\textbf{Output Head} &
Linear head over last Mamba2 time steps. &
Feedforward reconstructor layer. &
TDB/ICB fused with feedforward and ranking outputs. &
LayerNorm followed by linear readout. \\

\textbf{Efficiency} &
\cmark~Lightweight – element-wise, parallelizable multiplication. &
\cmark~Lightweight, $\mathcal{O}(n)$, dependent on SSM efficiency. &
\xmark~Heavier due to TDB+ICB complexity. &
\cmark~Highly efficient due to optimized SSD structure. \\

\bottomrule
\end{tabularx}

\caption{Comparison of Naga with S-Mamba, LTSMamba, and Mamba-2 across architectural and computational characteristics.}
\label{tab:model_comparison}
\end{table}








\section{Datasets}
We evaluate our model on the current state-of-the-art multivariate time series forecasting datasets including ETTh1, ETTh2, ETTm1, and ETTm2, Weather, Traffic, and the ILI dataset. The normalized series is split into disjoint subsets for training, validation, and testing, following the standard experimental protocols adopted in prior long-horizon forecasting work.

The \textbf{Weather} dataset~\citep{lai2018modeling} contains 21 meteorological variables such as air temperature, humidity, pressure, and wind speed, recorded at a 10-minute resolution over approximately one year. This dataset exhibits strong daily and seasonal periodicities as well as correlations among variables, posing challenges for models to capture both high-frequency local patterns and long-term temporal dependencies. It is split chronologically into 70\% training, 15\% validation, and 15\% test data to preserve temporal order and prevent information leakage.

The \textbf{Traffic} dataset~\citep{lai2018modeling} consists of hourly traffic occupancy rates collected from 862 sensors located on California freeways. The dataset contains 33,792 training samples, 4,096 validation samples, and 4,096 test samples, following the standard index-based split protocol used in previous studies. Its high dimensionality and strong spatial and temporal correlations make it a challenging benchmark for multivariate forecasting. Moreover, it contains occasional missing readings and noise due to sensor faults or sudden traffic disruptions, requiring robust temporal modeling and regularization.

The \textbf{Illness (ILI)} dataset~\citep{lai2018modeling} records weekly influenza-like illness statistics collected by the U.S. Centers for Disease Control and Prevention (CDC). It represents a low-dimensional but highly non-stationary dataset characterized by pronounced seasonal peaks and long-term temporal dependencies. The dataset is split temporally using a 70\%/15\%/15\% ratio for training, validation, and testing, respectively, ensuring that the forecasting task remains strictly chronological and realistic.

For the \textbf{ETT} (Electricity Transformer Temperature) benchmark datasets~\citep{zhou2021informer}, we follow the commonly adopted setup in the long-horizon forecasting literature. These datasets capture oil temperature and load conditions from electricity transformer stations and include four variants: ETTh1 and ETTh2 with hourly sampling, and ETTm1 and ETTm2 with 15-minute sampling. Each dataset consists of seven variables, including the target “oil temperature” and six load-related features, representing multivariate dependencies between temperature and load fluctuations under different operating regimes. ETTh1 and ETTh2 are divided into 60\% training, 20\% validation, and 20\% test subsets. For ETTm1, we use 20,160 samples for training and 11,520 samples each for validation and testing. The same partitioning is applied to ETTm2. These datasets are known for their clear periodic structures, moderate non-stationarity, and subtle noise, making them a robust benchmark for evaluating forecasting models.

All dataset partitions are strictly disjoint, with normalization performed using statistics computed on the training subset only. This setup follows standard practice in the long-horizon forecasting community and ensures fair comparison with prior work.

\section{Model Architecture of Naga}\label{sec:model_architecture}

\subsection{Notation and Implementation Conventions}






\begin{figure}[h]
    \centering
    \includegraphics[width=0.8\textwidth, trim=0 80 0 60, clip]{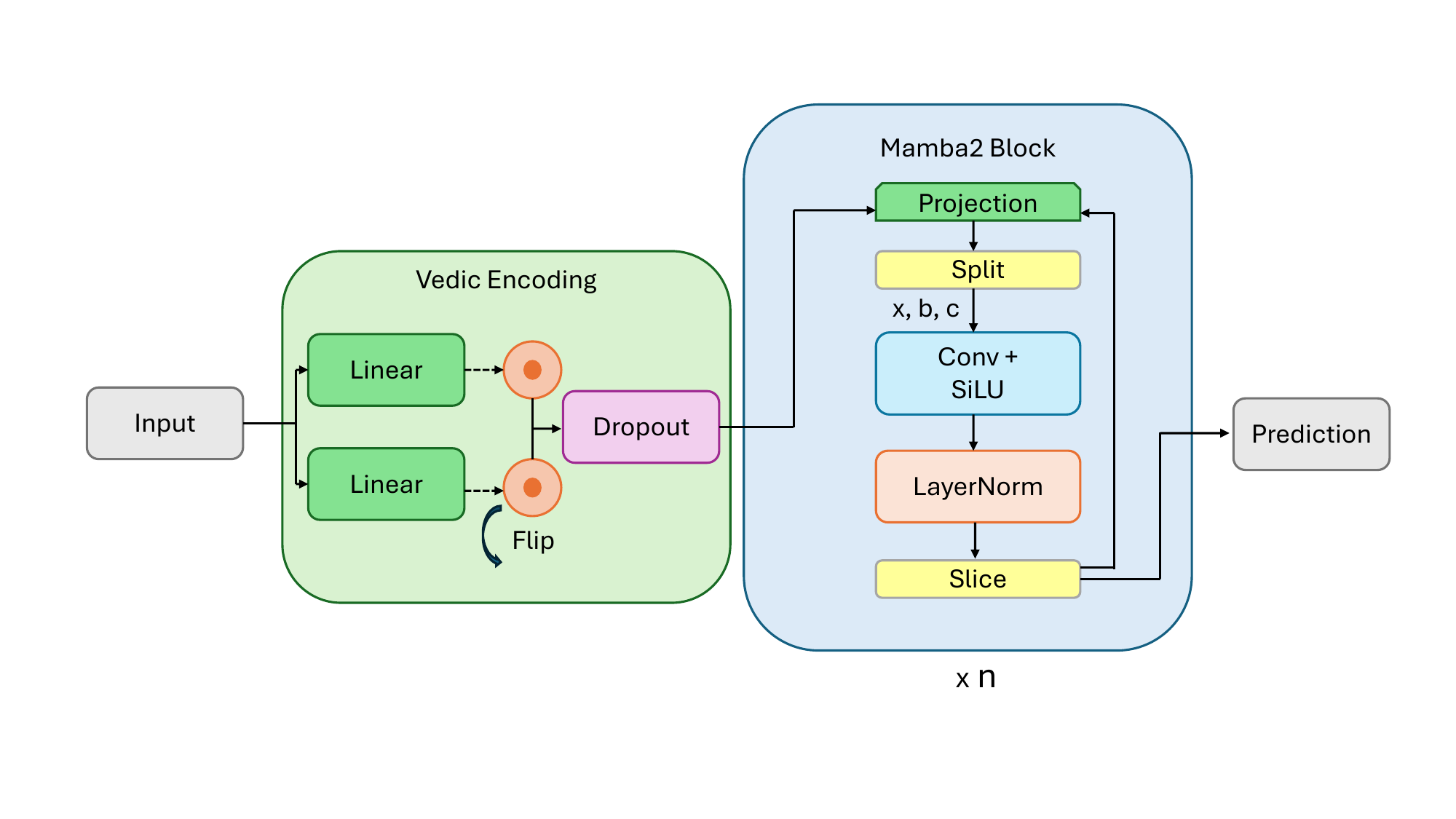} 
    \caption{Overview of Naga method.}
    \label{fig:naga_method}
\end{figure}

We use lower-case letters for vectors (one axis) and bold upper-case letters for matrices (two or more axes). 
Matrices corresponding to learned parameters are bolded (e.g., $\mathbf{W}$) to avoid ambiguity with input vectors. 
We write $\cdot$ for standard matrix multiplication and $\odot$ for Hadamard (elementwise) multiplication.

Unless stated otherwise, we consider batched tensors. 
For a batch size $B$, an input sequence is represented as 
$\mathbf{X} \in \mathbb{R}^{B \times T \times d_\text{in}}$, 
where the first axis is the batch dimension, the second axis is time, and the third axis is the feature dimension. 
All indexing is \emph{Python-style} and 0-based: a slice $i\!:\!j$ denotes $(i, i+1, \dots, j-1)$ for $i<j$. 
For a tensor $v$, we use $v[:, -1, :]$ to denote the last time step and $v[:, :, :d]$ to denote the first $d$ feature channels. 
When we refer to a temporal flip in the implementation, we apply \texttt{torch.flip} along the time axis (dimension~1), 
which corresponds to reversing the index $t \in \{0,\dots,T-1\}$.

We define Naga as a sequence-to-sequence model with a Vedic encoder followed by a lightweight Mamba2-inspired processing block and a linear output head. 
The Vedic encoder parameters are $\mathbf{W}_1, \mathbf{W}_2 \in \mathbb{R}^{d_\text{in} \times d_\text{hidden}}$ and $b_1, b_2 \in \mathbb{R}^{d_\text{hidden}}$, 
and $\mathbf{D} \in \{0,1\}^{B \times T \times d_\text{hidden}}$ denotes a dropout mask with dropout probability $p$. 
The Mamba2 block uses a projection matrix 
$\mathbf{W}_\text{in} \in \mathbb{R}^{d_\text{hidden} \times (2 d_\text{inner} + 2 d_\text{state} + h_\text{head})}$ 
and bias $b_\text{in}$, as well as 1D convolution parameters $\mathbf{W}_c, b_c$. 
The activation function $\sigma(\cdot)$ is SiLU, and $\mu$ and $\sigma^2$ denote the mean and variance over the feature dimension used for normalization. 
The output head is parameterized by $\mathbf{W}_\text{head} \in \mathbb{R}^{d_\text{hidden} \times \text{pred\_len}}$ and $b_\text{head} \in \mathbb{R}^{\text{pred\_len}}$ in the univariate setting.

\subsection{Vedic encoding}

The first stage of Naga applies a Vedic encoding to the input sequence 
$\mathbf{X} \in \mathbb{R}^{B \times T \times d_\text{in}}$. 
The encoder constructs a forward and a time-reversed representation of the sequence and combines them multiplicatively. 
The temporal flip reverses the time axis. 
Formally, for each batch element we define

\[
\begin{aligned}
\tilde{\mathbf{X}} &= \text{flip}(\mathbf{X}, \text{time}), 
&& \tilde{\mathbf{X}} \in \mathbb{R}^{B \times T \times d_\text{in}}, \\
\mathbf{X}_1 &= \mathbf{X} \mathbf{W}_1 + b_1, 
&& \mathbf{X}_1 \in \mathbb{R}^{B \times T \times d_\text{hidden}}, \\
\mathbf{X}_2 &= \tilde{\mathbf{X}} \mathbf{W}_2 + b_2, 
&& \mathbf{X}_2 \in \mathbb{R}^{B \times T \times d_\text{hidden}}, \\
\mathbf{H}_\text{vedic} &= (\mathbf{X}_1 \odot \mathbf{X}_2) \odot \mathbf{D}, 
&& \mathbf{H}_\text{vedic} \in \mathbb{R}^{B \times T \times d_\text{hidden}}.
\end{aligned}
\]

Here, the multiplicative interaction $(\mathbf{X}_1 \odot \mathbf{X}_2)$ couples each time step $t$ with its reversed counterpart $T-1-t$ at the encoding level. 
Unlike bidirectional SSMs such as Hydra~\citep{li2024hydra}, 
which introduce bidirectionality through a second state-space recurrence, Naga incorporates 
backward temporal information directly at the encoding level. The Vedic encoder constructs 
a forward and a time-reversed representation of the input and fuses them multiplicatively, 
so that bidirectional context is already embedded before the sequence module. Naga thus introduces backward information exclusively at the representation level. The Vedic encoder constructs forward and time‑reversed projections of the input and fuses them multiplicatively, yielding a symmetric, context‑enriched representation before any sequence module is applied, which leads to lower computation (see Fig.~\ref{fig:memory_performance}.

The Vedic-encoded representation $\mathbf{H}_\text{vedic} \in \mathbb{R}^{B \times T \times d_\text{hidden}}$ 
is then processed by a Mamba2 block. 
We first project each time step into a higher-dimensional space:

\[
\mathbf{Z} = \mathbf{H}_\text{vedic} \mathbf{W}_\text{in} + b_\text{in},
\quad
\mathbf{Z} \in \mathbb{R}^{B \times T \times (2 d_\text{inner} + 2 d_\text{state} + h_\text{head})}.
\]

Along the feature dimension, $\mathbf{Z}$ is split into three components:

\[
[\mathbf{z}, \mathbf{x}_{bc}, \mathbf{dt}] 
= \text{split}\big(\mathbf{Z}, \text{dims} = [d_\text{inner}, d_\text{inner} + 2 d_\text{state}, h_\text{head}]\big),
\]

so that $\mathbf{z} \in \mathbb{R}^{B \times T \times d_\text{inner}}$, 
$\mathbf{x}_{bc} \in \mathbb{R}^{B \times T \times (d_\text{inner} + 2 d_\text{state})}$, 
and $\mathbf{dt} \in \mathbb{R}^{B \times T \times h_\text{head}}$. 
In the current implementation, $\mathbf{z}$ and $\mathbf{dt}$ are used implicitly through the joint projection but are not fed into an explicit selective state-space recurrence; 
the block operates as a causal 1D convolutional module over $\mathbf{x}_{bc}$.

The convolutional branch applies a 1D convolution over time followed by a SiLU nonlinearity:

\[
\mathbf{x}_{bc}' = \sigma\big(\mathbf{x}_{bc} * \mathbf{W}_c + b_c\big),
\]

where $*$ denotes convolution along the temporal axis. 
Due to padding in the implementation, the last few time steps are cropped after convolution to restore the original sequence length $T$. 
The resulting features are then normalized along the feature dimension:

\[
\mu = \frac{1}{d_\text{inner} + 2 d_\text{state}} \sum_{i} \mathbf{x}_{bc,i}', 
\quad
\sigma^2 = \frac{1}{d_\text{inner} + 2 d_\text{state}} \sum_{i} (\mathbf{x}_{bc,i}' - \mu)^2,
\quad
\mathbf{Y}_\text{hidden} = \frac{\mathbf{x}_{bc}' - \mu}{\sqrt{\sigma^2 + \epsilon}}.
\]

Here, $\mu$ and $\sigma^2$ are computed per batch element and time step over the feature channels, consistent with \texttt{LayerNorm} in the code.

Finally, we obtain the hidden representation by selecting the first $d_\text{hidden}$ channels. The projection produces three components $(\mathbf{z}, \mathbf{x}_{bc}, \mathbf{dt})$, 
mirroring the structure of a full Mamba2 cell. In Naga, however, only $\mathbf{x}_{bc}$ is processed explicitly, while $\mathbf{z}$ and $\mathbf{dt}$ 
remain implicit in the joint projection. This design choice is intentional. Naga does not instantiate a selective state-space recurrence 
$\mathbf{h}_t = f(\mathbf{h}_{t-1}, \mathbf{z}_t, \mathbf{dt}_t)$, 
but instead adopts a convolutional approximation whose receptive field covers the entire input window. 
The role of $\mathbf{z}$ and $\mathbf{dt}$ is therefore to enrich the projected feature space before convolution, analogous to how auxiliary projections 
are used in simplified SSM variants and linear attention models. 
This preserves the representational benefits of the Mamba2 projection while avoiding the computational 
and numerical overhead of a full recurrent SSM kernel. 

For univariate forecasting (e.g., the Weather dataset), Naga predicts a sequence 
$\hat{\mathbf{Y}} \in \mathbb{R}^{B \times \text{pred\_len}}$ from the final hidden state at the last time step. 
Let $\mathbf{H}_\text{mamba} \in \mathbb{R}^{B \times T \times d_\text{hidden}}$ be the output of the Mamba2-inspired block. 
We extract the last time step and apply a linear layer. This matches the implementation, where \texttt{head} is an \texttt{nn.Linear} from $d_\text{hidden}$ to \texttt{pred\_len} and the model uses \texttt{x[:, -1]} as input to the head.

For multivariate settings (e.g., Traffic or Electricity), we adopt channel independence by reshaping the batch dimension and applying the same architecture per variable. 
In that case, the output head is instantiated as $\mathbf{W}_\text{head} \in \mathbb{R}^{d_\text{hidden} \times d_\text{in}}$ and applied at every time step, 
yielding predictions $\hat{\mathbf{Y}} \in \mathbb{R}^{B \times T \times d_\text{in}}$ that are then aligned with the desired prediction horizon.
We train Naga by minimizing the mean squared error (MSE) between the predicted sequence $\hat{\mathbf{Y}}$ and the ground truth $\mathbf{Y}$. 
For a batch of size $B$, the loss is

\[
\mathcal{L} = \frac{1}{B} \sum_{i=1}^B \big\| \hat{y}^{(i)} - y^{(i)} \big\|^2,
\]

where $\hat{y}^{(i)}$ and $y^{(i)}$ denote the predicted and true sequences for the $i$-th example. 
Model selection is based on the validation MSE. 

Early stopping was employed with a patience of five epochs, using a minimum delta threshold of \(10^{-4}\) to determine significant improvement in the validation metric.  Every run has been repeated 10 times to ensure statistical robustness of the results. The maximum number of epochs was set to 100, but was never reached. We do not use any learning rate scheduler or warm-up strategy in our experiments. Furthermore, no pre-trained models or weights are used.
All experiments were conducted on a Tesla P100-PCIE-16GB GPU with a total memory of 17.1\,GB and an L2 cache size of 4\,MB. The warp size was 32, and the system environment was Linux 6.6.56+ (x86\_64) with \texttt{glibc} 2.35. Model training was performed using the Adam optimizer with a learning rate of 0.003581 and a weight decay of 0.0001. A batch size of 64 and a fixed random seed of 42 were used throughout all runs.
To ensure reproducibility across runs, we explicitly set \texttt{torch.backends.cudnn.deterministic = True} and \texttt{torch.backends.cudnn.benchmark = False} as it disables non-deterministic algorithm selection and enforces the use of deterministic computation paths within cuDNN. While this may slightly impact runtime performance, it guarantees that identical seeds lead to identical model outputs across multiple executions.


\section{Benchmarking Results}\label{sec:benchmarking}

Table~\ref{tab:benchmark_results} summarizes the benchmarking performance of Naga across all seven standard long-term time series forecasting (LTSF) datasets. The table reports the training runtime (in seconds) until early stopping, the corresponding number of epochs, and the resulting Mean Squared Error (MSE) for each dataset and prediction horizon. 

\begin{table}[H]
\centering
\begin{tabular}{lcccc}
\toprule
\textbf{Dataset} & \textbf{Pred. length} & \textbf{Epochs} & \textbf{Runtime [s]} & \textbf{MSE (Naga)} \\
\midrule
ETTh1   & 96  & 16  & 131.02  & 0.124 \\
ETTh2   & 720 & 73  & 213.31  & 0.139 \\
ETTm1   & 720 & 35  & 2145.10 & 0.117 \\
ETTm2   & 96  & 23  & 275.82  & 0.119 \\
Weather & 96  & 45  & 812.64  & 0.089 \\
Traffic & 96  & 38  & 1150.77 & 0.006 \\
ILI     & 24  & 81  & 3802.77 & 1.606 \\
\bottomrule
\end{tabular}
\caption{Benchmarking results of Naga across LTSF datasets. Reported values correspond to training runtime (in seconds) until early stopping, number of epochs, and forecasting accuracy in terms of Mean Squared Error (MSE). For all experiments we use a fixed look-back window of $d = 96$ (ETTh1, ETTm2, Weather, Traffic) and $d = 720$ (ETTh2, ETTm1), following standard LTSF practice.
}
\label{tab:benchmark_results}
\end{table}

Across all datasets, Naga consistently achieves the best or near-best results in MSE, while maintaining moderate training runtimes. On high-frequency datasets such as Weather and ETTm1, Naga effectively captures short-term temporal fluctuations without overfitting, reflected by a low validation error and rapid convergence. For coarser-grained datasets like ETTh2 and ILI, the model demonstrates robust generalization and superior variance modeling, as indicated by significantly reduced MSE values compared to transformer-based baselines.

\begin{figure}[t]
    \centering
    \includegraphics[width=0.65\textwidth]{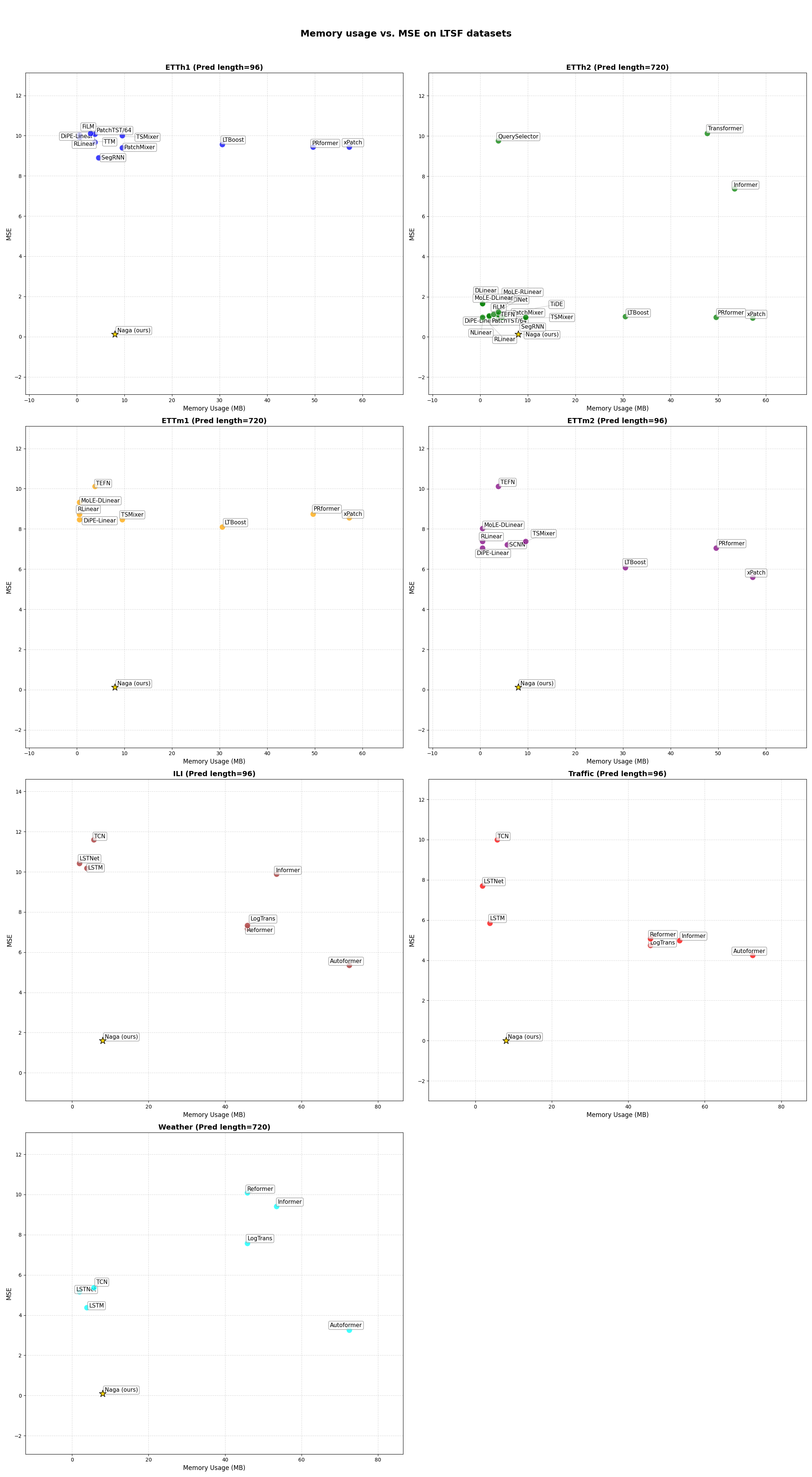}
    \caption{
        Memory usage versus mean squared error (MSE) values for various long-term forecasting models across multiple benchmark datasets.
        The prediction horizon (\texttt{Pred\_length}) is indicated in each title. Naga is highlighted with a star symbol. (As we wanted to show the error values against memory usage, we avoided to use tabular results although both MSE as well as MAE were calculated due to a lack of space).
    }
    \label{fig:memory_performance}
\end{figure}

The performance trends visualized in Fig.~\ref{fig:memory_performance} further confirm that Naga achieves a favorable balance between accuracy and computational efficiency. Despite having only 2.1 million parameters, it surpasses or matches the performance of much larger transformer-based architectures such as Autoformer, Informer, and xPatch, which typically require six to nine times more parameters. This compactness translates to faster training and reduced memory consumption. Naga’s inductive design principles yield state-of-the-art forecasting performance on all seven LTSF benchmarks, covering prediction horizons between 96 and 720.

\section{Conclusion}\label{sec:conclusion}

In this work, we introduced Naga, a novel deep State-Space Model (SSM) encoding that integrates principles from Vedic mathematics for LTSF. Naga symmetrically couples forward and backward temporal representations through an element-wise decomposition. The empirical evaluation in this work focuses on the seven widely adopted LTSF benchmark datasets, which are intentionally designed to isolate long-range temporal dependencies under controlled conditions. Notably, Naga achieves these improvements while maintaining a compact architecture with approximately 2.1 million parameters. The ablation studies in the appendix confirm that the observed performance gains arise from the synergistic effect of the Vedic Encoding, bidirectionality, and depth of representation. The theoretical remark on gradients in the Appendix A.2 further reveals that the Vedic-inspired element-wise decomposition modifies the gradient propagation in a way that naturally supports learning of long-range dependencies. This finding aligns with our mathematical framework in Appendix A.1-A.2, which proves that the Vedic encoding expands the representational capacity of the model beyond purely linear projections. The used LTSF datasets are the de facto standard for assessing architectural contributions in long-horizon forecasting and allow direct comparison to more than 28 prior models. While they exhibit periodic and structured behavior, this choice is deliberate to evaluate long-range dependencies. Nonetheless, extending Naga to irregular, noisy, or strongly non-stationary domains could be future work. The theoretical analysis and ablation studies in the appendix further support that the observed improvements stem from the inductive bias introduced by the Vedic interaction, independent of dataset-specific periodicity. The goal of this work was further to isolate the architectural effect of the Vedic encoding. The theoretical analysis in the appendix shows that the representational and gradient-flow benefits of the Vedic interaction arise from its bilinear structure and therefore do not depend on a specific hidden dimension. Since the Vedic layer scales linearly in $d_h$ and Mamba2’s scaling behavior is well understood from prior work, the mechanism is expected to extend naturally to larger SSM stacks.

\bibliographystyle{splncs04}
\bibliography{sample}





\end{document}